# TGLD: A Trust-Aware Game-Theoretic Lane-Changing Decision Framework for Automated Vehicles in Heterogeneous Traffic

Jie Pan[1], Tianyi Wang[2†], Yangyang Wang[3], Junfeng Jiao[4], and Christian Claudel[2]

***Abstract*—Automated vehicles (AVs) face a critical need to adopt socially compatible behaviors and cooperate effectively with human-driven vehicles (HVs) in heterogeneous traffic environment. However, most existing lane-changing frameworks overlook HVs' dynamic trust levels, limiting their ability to accurately predict human driver behaviors. To address this gap, this study proposes a trust-aware game-theoretic lane-changing decision (TGLD) framework. First, we formulate a multi-vehicle coalition game, incorporating fully cooperative interactions among AVs and partially cooperative behaviors from HVs informed by real-time trust evaluations. Second, we develop an online trust evaluation method to dynamically estimate HVs' trust levels during lane-changing interactions, guiding AVs to select context-appropriate cooperative maneuvers. Lastly, social compatibility objectives are considered by minimizing disruption to surrounding vehicles and enhancing the predictability of AV behaviors, thereby ensuring human-friendly and context-adaptive lane-changing strategies. A human-in-the-loop experiment conducted in a highway on-ramp merging scenario validates our TGLD approach. Results show that AVs can effectively adjust strategies according to different HVs' trust levels and driving styles. Moreover, incorporating a trust mechanism significantly improves lane-changing efficiency, maintains safety, and contributes to transparent and adaptive AV-HV interactions.**

## I. INTRODUCTION

Despite rapid advancements, fully automated driving still faces substantial challenges that limit its widespread adoption in real-world traffics [1]. As a result, heterogeneous traffic environments, where human-driven vehicles (HVs) and automated vehicles (AVs) coexist, will remain common for the foreseeable future [2]. HVs rely heavily on experience, intuition, and non-verbal communication, whereas AVs operate depending on sensor data, perception algorithms, and rule-based logics, making them susceptible to misinterpreting human intentions [3]. Additionally, many drivers exhibit low trust in AVs, especially in bottlenecks like highway on-ramp merging areas, due to concerns over safety, privacy, and cyber threats [4]. This lack of trust can lead to uncertain human responses, increased interaction conflicts, and ultimately degrade traffic efficiency and safety. Therefore, AVs must be capable of dynamically adapting their decision-making strategies based on HVs' trust levels and driving styles.

Trust has been extensively studied across fields such as psychology, human-computer interaction, and automation [5]. In automation, trust refers to the user's willingness to rely on an automated system, influenced by factors including the user's knowledge, gender, and age [6]. Research in human-robot interaction indicates that trust is defined as a dynamic, context-dependent variable shaped by the system reliability, predictability, and transparency [7]. Initial research mainly measured and modeled trust, while identifying factors influencing trust in AVs, such as its impact on takeover behavior in semi-autonomous driving [8]. Recent studies have moved toward enabling AVs to infer and shape human trust levels actively. For example, trust has been modeled as a hidden variable in deep reinforcement learning frameworks to guide robots based on inferred trust from observed behavior [9]. Others have incorporated trust mechanism into multi-agent reinforcement learning to regulate inter-agent cooperation [10]. However, these models typically view trust as a static or linear variable, whereas practical cases reveal that trust evolves dynamically through past interactions, expectations, and real-time behavioral feedback, particularly in collaborative tasks. Therefore, relying solely on trust propensity fails to capture its dynamic nature in HV-AV interactions. Additionally, most existing frameworks primarily address homogeneous scenarios, limiting their applicability in heterogeneous environments, and often treat trust as an implicit factor rather than an explicit design requirement.

Currently, significant progress has been made in lane-changing decision models, including rule-based [11], optimization-based [12], learning-based [13], and game theory-based [14] approaches. For instance, Zhao et al. [15] employed rough set theory to extract lane-changing rules from naturalistic driving data, achieving practical effectiveness. However, rule-based approaches struggle in handling dynamic or unexpected scenarios. Optimization-based methods can enhance flexibility by tuning parameters in response to varying traffic conditions. Wiesner et al. [16] demonstrated the effectiveness of particle swarm optimization in improving decision-making accuracy and efficiency, though such methods are heavily data-intensive and computationally demanding. Game theory-based models are particularly well-suited for heterogeneous traffic, as they can capture the strategic interactions between HVs and AVs. Yu et al. [17] developed a predictive game-theoretic controller to enhance safety in mandatory lane changes in mixed traffic. Guo et al. [18] integrated driver risk perception into a Nash equilibrium framework to examine steering interactions in shared control systems. Integrating game theory with deep Q networks and proximal policy optimization has further improved the adaptability and efficiency of these models. However, current research predominantly focuses on cooperative AV environments, neglecting the complexities and variability inherent in human driving behaviors, including driving styles, risk perception, and especially trust towards AVs. These human-centric factors are essential for predicting accurate behavior and achieving socially-aware AV decision-making in heterogeneous traffic environments.

†Corresponding author: Tianyi Wang.

[1]Jie Pan is with the Department of Civil Engineering, Tsinghua University, Beijing 100084, China. Email: panj21@mails.tsinghua.edu.cn.

[2]Tianyi Wang and Christian Claudel are with the Department of Civil, Architectural, and Environmental Engineering, The University of Texas at Austin, Austin, TX 78712, USA. Email: bonny.wang@utexas.edu; christian.claudel@utexas.edu.

[3]Yangyang Wang is with the School of Automotive Studies, Tongji University, Shanghai 201804, China. Email: wyangyang@tongji.edu.cn.

[4]Junfeng Jiao is with the School of Architecture, The University of Texas at Austin, Austin, TX 78712, USA. Email: jjiao@austin.utexas.edu.

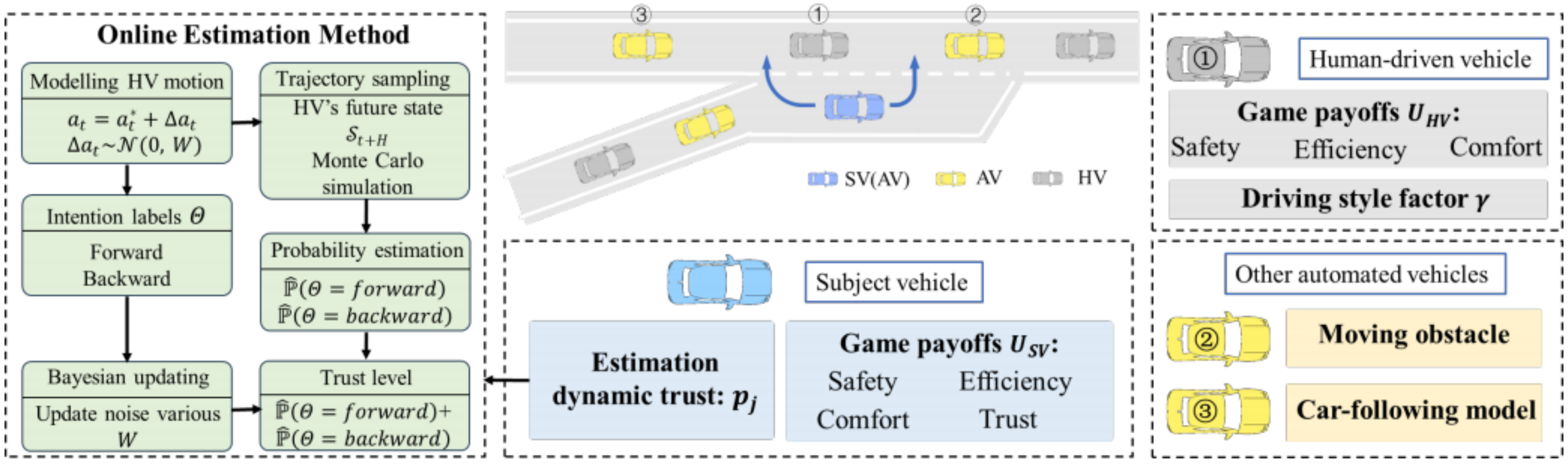

Fig. 1. Framework of the proposed trust-aware game-theoretic lane-changing decision (TGLD) framework

To address the challenges outlined above, this study proposes a trust-aware game-theoretic lane-changing decision (TGLD) framework for AVs in heterogeneous traffic environments, as shown in Figure 1. The main contributions are as follows:

- A trust-aware multi-vehicle coalition game in heterogeneous traffic is proposed, including a fully cooperative game among AVs and a partial cooperative game for HVs due to different trust levels.
- An online trust estimation method is developed to continuously update trust levels by leveraging both recent behavioral interactions and long-term driving style characteristics.
- Social compatibility objectives are considered in AVs' payoff function, aiming to minimize disruption to surrounding vehicles and enhance the predictability of AV behavior, thereby promoting trustworthy, and socially acceptable driving behaviors.

## II. METHODOLOGY

### A. Trust-Aware Multi-Vehicle Coalition Game

This study focuses on the highway on-ramp merging scenario as shown in Figure 1, where a heterogeneous traffic environment exists with both AVs and HVs on the mainline and the on-ramp. To achieve safe and efficient merging, the ideal behavior involves mainline vehicles proactively accelerating or decelerating to create sufficient gaps, while on-ramp vehicles adjust their speeds accordingly and merge at an opportune moment, thus avoiding collisions or deadlock situations. However, the introduction of HVs significantly complicates the coordination due to uncertainty in behavior and varying trust levels toward AVs. In this context, the subject vehicle (SV) is an AV attempting to merge, which must determine the optimal lane-changing timing and execute the merging maneuver based on the types of vehicles encountered and a real-time trust evaluation. Consequently, this merging task involves both decision-making and trajectory-planning.

In an n-player game, any non-empty subset $\Omega$ of the set $N$ consisting of all participants can be defined as a coalition. In the collaborative region shown in Figure 1, the set of all vehicles is denoted as $N = \{V_1, \cdots, V_n\}$, where $V_n$ denotes the n-th vehicle, $\Omega$ denotes the coalition formed by $V_i$, i.e. $\Omega \in N$, $\Omega_{AV}$ denotes the set $\Omega$ of all AVs in the coalition, and $\Omega_{HV}$ denotes the set of all HVs. In coalition, AVs share information and reach the following agreements: (1) when two AVs have a lane-changing conflict, a cooperative game strategy is used; (2) if an HV has a high level of trust in an AV, the cooperative behavior needs to be considered. To demonstrate the rationality of the coalition agreement, let $v$ be the characteristic function corresponding to each coalition in the set $N$, where $v(\Omega)$ represents the total benefit that all participants in coalition $\Omega$ can obtain:

$$v(\Omega) = \sum_{i \in \Omega_{AV}} u_i + \sum_{j \in \Omega_{HV}} (p_j \cdot u_j) \tag{1}$$

where $p_j \in [0,1]$ denotes the trust value of HV towards AV, which is used to adjust the HV's contribution to the overall coalition payoff; and let $U = \{u_1, \cdots, u_m\}$ be an m-dimensional vector, where each $u_i (i = 1, 2, \cdots, m)$ denotes the gain allocated to $V_i$ in coalition.

If the coalition game satisfies:

$$u_i \geq v(\Omega / \{V_i\}) \tag{2}$$

then the game meets the **individual rationality** condition, where $v(\Omega/\{V_i\})$ represents the payoff if $V_i$ does not participate in the coalition.

If the coalition game satisfies:

$$\sum_{i=1}^{m} u_i = v(\Omega) \tag{3}$$

then the game satisfies the **group rationality** condition.

Only when the coalition agreement satisfies both the **group rationality** and **individual rationality** conditions can an AV accept the agreement. In such cases, the overall coalition payoff is defined as the sum of the individual payoffs of the vehicles, thereby satisfying the **group rationality** condition.

When multiple vehicles encounter conflicts within the cooperation area, the coalition game can be transformed into multiple two-vehicle lane-changing sub-problems. If both vehicles involved in a lane-changing game are AVs, a **cooperative game** strategy is employed. However, if the interaction involves an HV and an AV, a **non-cooperative game** strategy is adopted. The primary distinction between cooperative and non-cooperative games is the presence of binding agreements prior to the commencement of the game. In **cooperative games**, participants have the ability to establish enforceable agreements, such as those concerning the distribution of payoffs. In contrast, **non-cooperative games** lack such binding agreements, requiring each agent to act based on its own strategy without enforceable commitments.

### *B. Human-Driven Vehicles with Different Driving Styles*

The human driver's control preferences and driving styles are reflected in their observable driving behaviors. Building upon the payoff functions in [19], this study incorporates the driving style factor [14] to describe a driver's individual preference for aggressiveness. Thus, accurate modeling of driver payoffs is essential for the SV to make rational and context-aware decisions. To evaluate the safety payoff $U_{safety}^{FV}$, the time headway (THW) is used. Jerk is used to quantify the comfort payoff $U_{comfort}^{FV}$. Traffic efficiency $U_{eff}^{FV}$ is assessed by the absolute deviation between the actual speed $v_{FV}$ and the desired speed $v_{desire}$. Then, the total payoff function $U_{FV}$ is designed to reflect different drivers:

$$\begin{aligned} U_{FV} =& (1-\gamma)\ \lambda_{safe} U_{safe}^{FV} + \gamma \lambda_{eff} U_{eff}^{FV} \\ &+ (1-|1-2\gamma|)\ \lambda_{com} U_{com}^{FV} \end{aligned} \tag{4}$$

where $\lambda_{safe}$, $\lambda_{com}$, and $\lambda_{eff}$ are the weight coefficients for safety, comfort, and efficiency payoffs, respectively; and $\gamma \in [0,1]$ is the FV's driving style factor. In this study, the parameter $\gamma$ is set to 0.7, 0.5, and 0.3, corresponding to aggressive, normal, and defensive driving styles, respectively.

In heterogeneous traffic environments, it is essential for AVs to recognize the driving styles of surrounding HVs prior to interaction. This study extends the approach proposed by [20] to develop a driving style classification model based on trajectory data. Rather than attempting to comprehensively analyze the psychological origins or underlying mechanisms behind individual driving styles, the emphasis is placed on understanding their observable impact on driving behavior. This enables AVs to improve the accuracy of behavior prediction and rationality of lane-changing decisions, thereby facilitating safer and more socially compatible interactions.

### *C. Automated Vehicles with Different Trust Levels*

In addition to the previously introduced payoff components including safety, efficiency, and comfort, trust plays a critical role in AV decision-making. This component encourages the AV to act in ways that are perceived as trustworthy by surrounding human drivers. Specifically, the trust reward is formulated as a weighted sum of two key influencing factors:

$$U_{trust}^{SV} = \xi_1 U_{trust,1}^{SV} + \xi_2 U_{trust,2}^{SV} \tag{5}$$

*1) Minimizing Impact on Others $U_{trust,1}^{SV}$:* This term encourages the SV to minimize negative effects on surrounding vehicles, such as avoiding to brake sharply during lane change. It is modeled as the probability that the FV chooses a favorable response $(a^{FV}, c^{FV})$ given the current state $S$ and the SV's action $(a^{SV}, c^{SV})$. This formulation reflects how likely the FV is to respond positively, assuming rational behavior guided by their payoff function $U_{FV}$.

$$\begin{aligned} U_{trust,1}^{SV} =& P\left((a^{FV}, c^{FV}) | S, (a^{SV}, c^{SV})\right) \\ =& \frac{e^{U_{FV}\left((a^{FV}, c^{FV}) | S, (a^{SV}, c^{SV})\right)}}{\sum_{\forall (a^{FV}, c^{FV})} e^{U_{FV}\left((a^{FV}, c^{FV}) | S, (a^{SV}, c^{SV})\right)}} \end{aligned} \tag{6}$$

*2) Enhancing Self-Predictability $U_{trust,2}^{SV}$:* This term encourages the SV to make its behavior more predictable, thus reducing uncertainty for surrounding vehicles. Predictable behavior may include decisive lane changes, minimal hesitation, and consistent motion. It is defined as the difference between the probabilities of the SV's most and second-most likely actions, conditioned on the FV's response:

$$\begin{aligned} U_{trust,2}^{SV} =& P\left((a_{first}^{SV}, c_{first}^{SV}) | S, (a^{FV}, c^{FV})\right) \\ &- P\left((a_{second}^{SV}, c_{second}^{SV}) | S, (a^{FV}, c^{FV})\right) \end{aligned} \tag{7}$$

A larger difference indicates a more predictable choice by the AV. Therefore, the total payoff for the SV is:

$$U_{SV} = k_{safe} U_{safe}^{SV} + k_{eff} U_{eff}^{SV} + k_{com} U_{com}^{SV} + k_{trust} U_{trust}^{SV} \tag{8}$$

### *D. Online Estimation Method for Dynamic Trust*

In high-density traffic, the lane-changing maneuvers of vehicles are often constrained, making acceleration the most observable and informative behavior variable. This study focuses on estimating HV trust dynamically based on longitudinal motion. The HV's action at time $t$ is modeled as:

$$a_t = a_t^* + \Delta a_t, \Delta a_t \sim N(0, W) \tag{9}$$

where $a_t^*$ is the expected action, and $\Delta a_t$ is behavioral uncertainty. Using an LSTM-based prediction model, the HV's future state $S_{t+H}$ is derived recursively:

$$S_{t+H} = f(a_{t+1}, a_{t+2}, \cdots, a_{t+H-1} | S_t) \tag{10}$$

Due to uncertainty in future actions, Monte Carlo sampling is applied to estimate the trajectory distribution. For each of the $Z$ sampled trajectories, intention labels $\Theta \in \{forward,\ backward,\ uncertain\}$ are assigned based on position and velocity thresholds $(\delta_x, \delta_v, \varepsilon_x, \varepsilon_v)$:

$$\mathscr{F} = \{x_k^{HV},\ v_k^{HV} | (\Delta x_k > \delta_x) \cup (\Delta v_k > \delta_v)\} \tag{11}$$

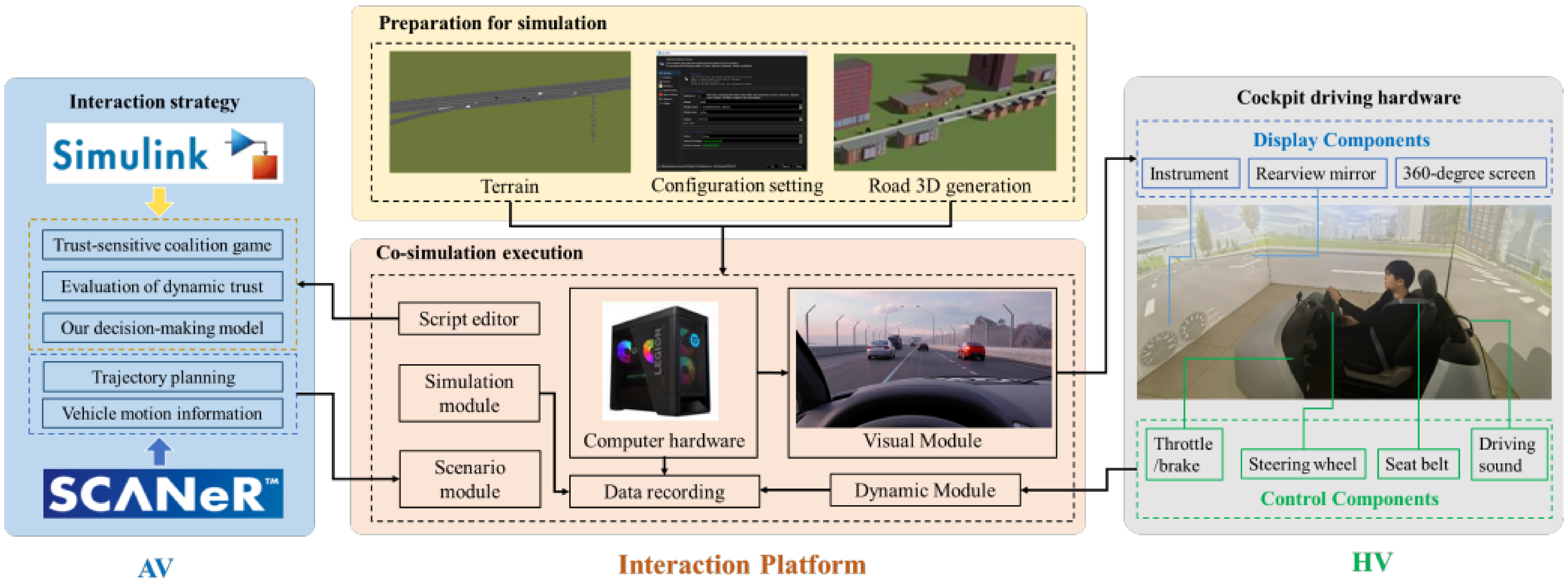

Fig. 2. Human-in-the-loop (HIL) driving platform

$$\mathscr{B} = \{x_k^{HV},\ v_k^{HV} | (\Delta x_k < -\varepsilon_x) \cup (\Delta v_k < -\varepsilon_v)\} \tag{12}$$

Trajectory intention probabilities are estimated by counting occurrences:

$$\hat{\mathbb{P}}(\Theta = forward) = \frac{1}{Z}\sum_{z=1} \chi(\Theta^{(z)} = forward) \tag{13}$$

$$\hat{\mathbb{P}}(\Theta = backward) = \frac{1}{Z}\sum_{z=1} \chi(\Theta^{(z)} = backward) \tag{14}$$

Then, HV's trust level $p_j$ is computed as:

$$\begin{aligned} p_j =& \hat{\mathbb{P}}(\Theta \neq uncertain) \\ =& \hat{\mathbb{P}}(\Theta = forward) + \hat{\mathbb{P}}(\Theta = backward) \end{aligned} \tag{15}$$

where, $p_j$ is initially set at 0.5 and is dynamically updated. To capture behavioral uncertainty, the noise variance $W$ is updated online using Bayesian inference.

## III. EXPERIMENTS

### A. Experimental Setup

To validate the performance of the proposed TGLD framework, a human-in-the-loop (HIL) experiment is conducted using a 360-degree surround-screen driving simulator at Tsinghua University. Following the methodology in [8], participants' driving styles are classified using the aggressive driving scale. A total of 36 drivers are recruited in the study, including 18 aggressive and 18 defensive drivers. These participants have an average of 5.4 years of driving experience and report annual driving mileage between 1,500 and 30,000 kilometers. As illustrated in Figure 2, the experiment utilizes a high-fidelity virtual environment via SCANeR Studio. The decision-making and trust estimation algorithms are implemented in Matlab/Simulink. A real-time synchronization module is also deployed in Simulink to ensure that the simulation runs in sync with real-world time.

When the SV merges into the highway mainline, it must consider vehicles in both the current lane and the target lane, and choose an appropriate gap based on the cooperation willingness of surrounding vehicles. This merging scenario can be simplified in Figure 1, where the SV represents an AV initiating a lane change and other yellow vehicles (AVs) are capable of communicating with the SV in advance and participating in cooperative lane-changing. HV① is a human-driven vehicle, while AV② is an automated vehicle that, due to its leading position, does not interact with the SV. In the experiment setup, all vehicles, except HV①, are initialized with a speed of 15 m/s. The initial distance between AV③ and AV② is set to 50 meters, and the SV's perception range is set to 150 meters. Once detecting an upcoming merging, the SV must determine the appropriate timing and select a suitable gap to merge into the left lane, ensuring safety, comfort, and efficiency.

During the experiment, drivers are instructed to drive according to their natural driving style. When the SV initiates a lane-changing intention by activating the left-turn signal, the HV could decide whether to yield or not based on the driver's own judgment. After completing each lane-changing event, the driver is instructed to pull over and complete a trust in automation questionnaire [5]. According to the collected questionnaire results, the experiments are classified into 21 cases of high trust and 15 cases of low trust.

### B. Trajectory Analysis

To demonstrate the feasibility and real-time performance of the proposed TGLD framework, four representative experimental cases are selected from the HIL simulation data. These cases represent various combinations of driving styles (defensive vs. aggressive) and trust levels (high vs. low). Trajectories are plotted from the moment the SV initiates a lane-changing intention. As shown in Figure 3, orange, blue, purple, and green rectangles represent the SV, HV, other AV, and the leading vehicle (LV), respectively. Solid rectangles denote vehicle positions at 0 s, while dashed ones indicate positions at 8 s and 13 s.

### C. Comparative Performance Analysis

We evaluate the proposed TGLD framework along three dimensions: safety, efficiency, and comfort. **Safety** is as-sessed by the *minimum THW* and *minimum distance* between the SV and HV during lane changes; **efficiency** by the *average speed*; and **comfort** by the *maximum jerk* experienced.

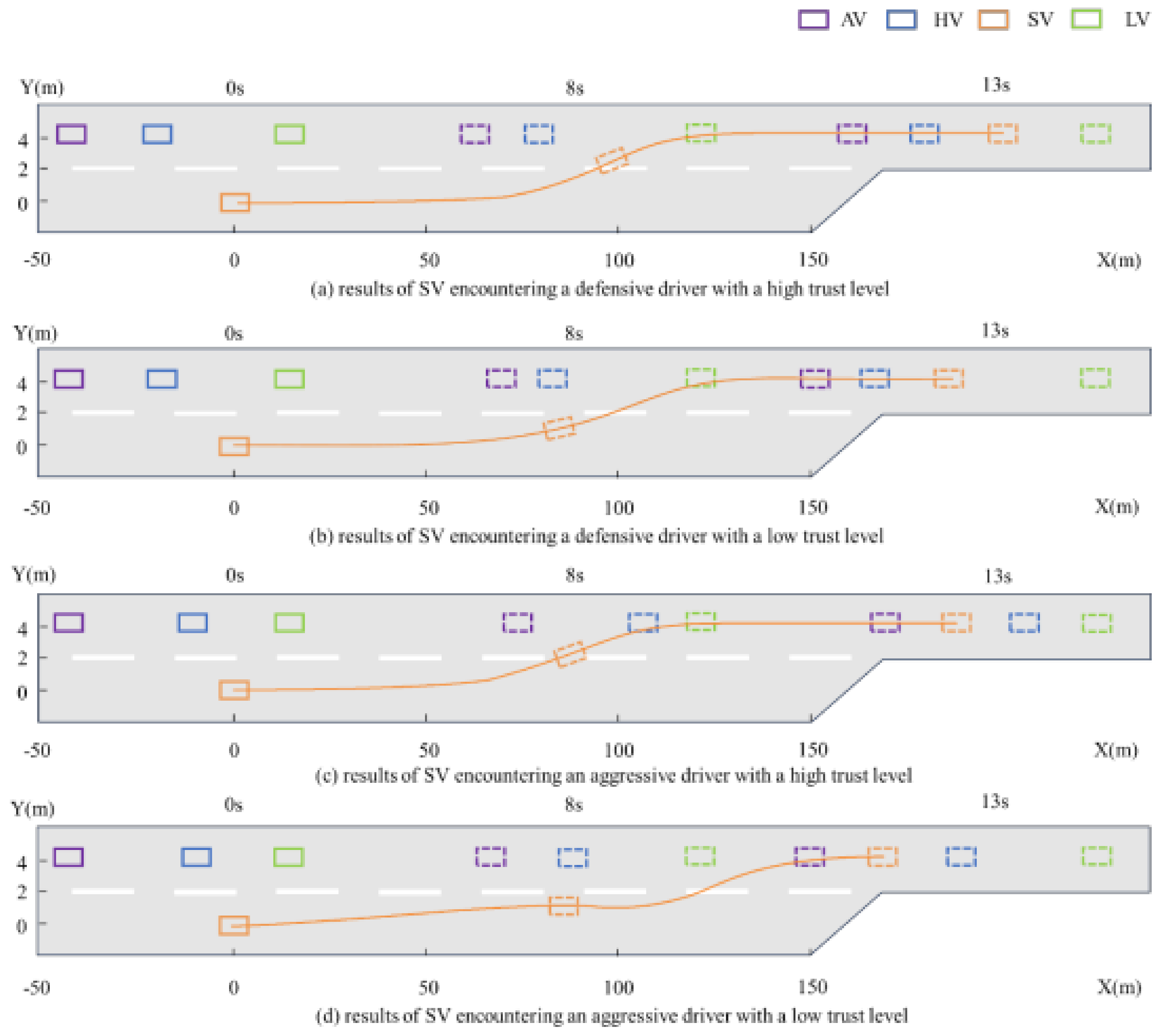

Fig. 3. Lane-changing trajectories of the subject vehicle (SV)

As shown in Figure 4, defensive drivers with high trust (D-H) maintain the highest **safety** margins, with medians of 2.44 s and 20.94 m and no low outliers, indicating safer interactions. In the defensive drivers with low trust (D-L) condition, greater variability appears, with *minimum THW* dropping to 0.84 s and *minimum distance* to 13 m, reflecting inconsistent, unstable reactions. For aggressive drivers, the aggressive drivers with high trust (A-H) condition shows relatively high and concentrated safety margins, while aggressive drivers with low trust (A-L) displays generally lower values, although still safer than defensive low-trust cases, as AVs tend to decelerate to merge safely.

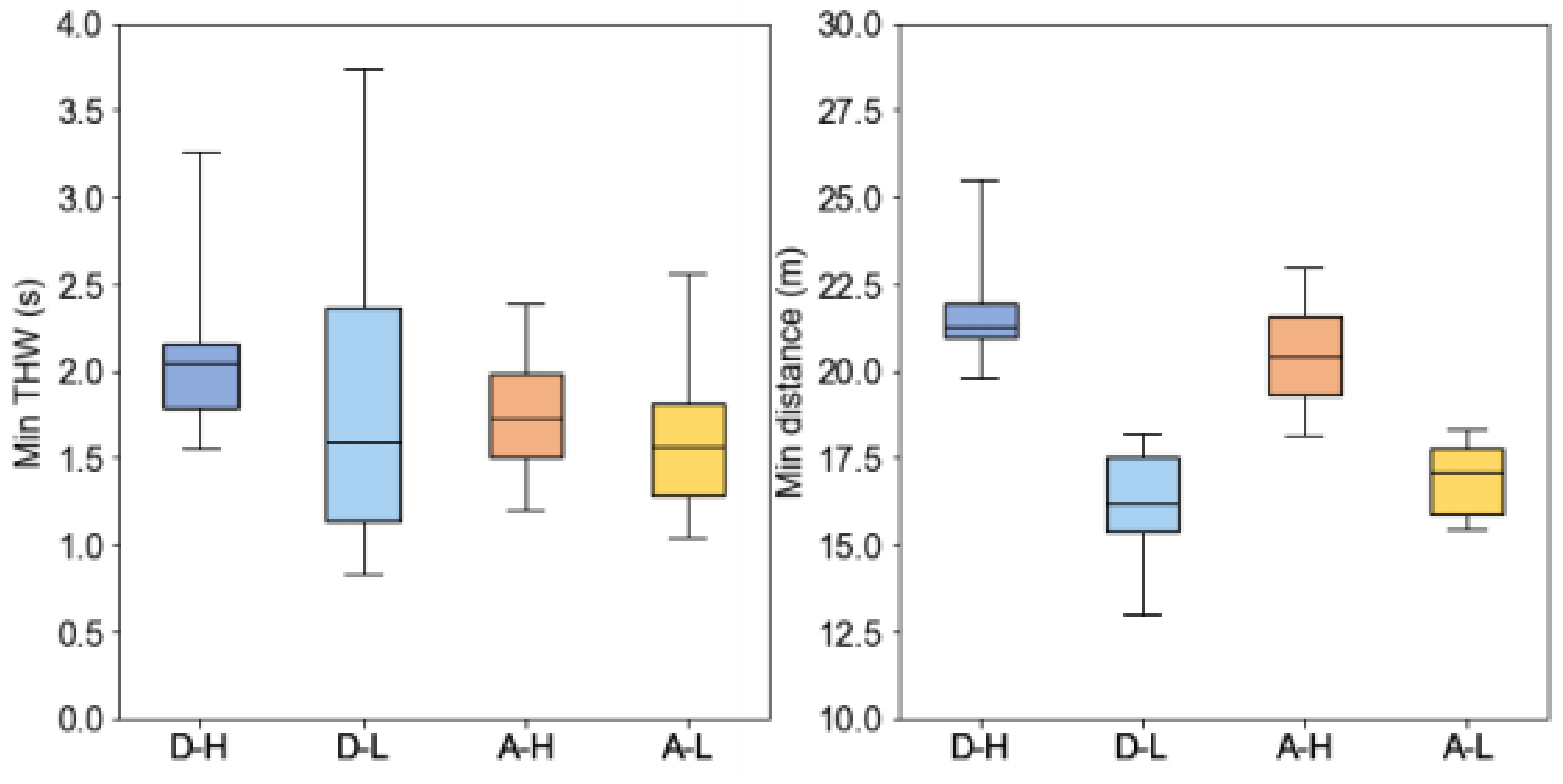

Fig. 4. Comparison of safety evaluation indicators

**Efficiency** results in Figure 5 show that D-H yields moderate and stable *average speeds* (HV: 13.11 m/s; SV: 15.1 m/s), while D-L sees lower and more variable HV speeds, reducing flow stability. In A-H, HV speeds are higher and more concentrated (15.51–17.58 m/s), supporting efficient merges. The SV slightly reduces its speed (14.51 m/s) to yield space, reflecting an efficient and coordinated merging process. In the A-L condition, the HV displays significant speed variation due to distrust, and the SV's average speed decreases markedly. This demonstrates that low trust from aggressive drivers forces SV to adopt more defensive strategies, increasing speed differentials and reducing traffic efficiency.

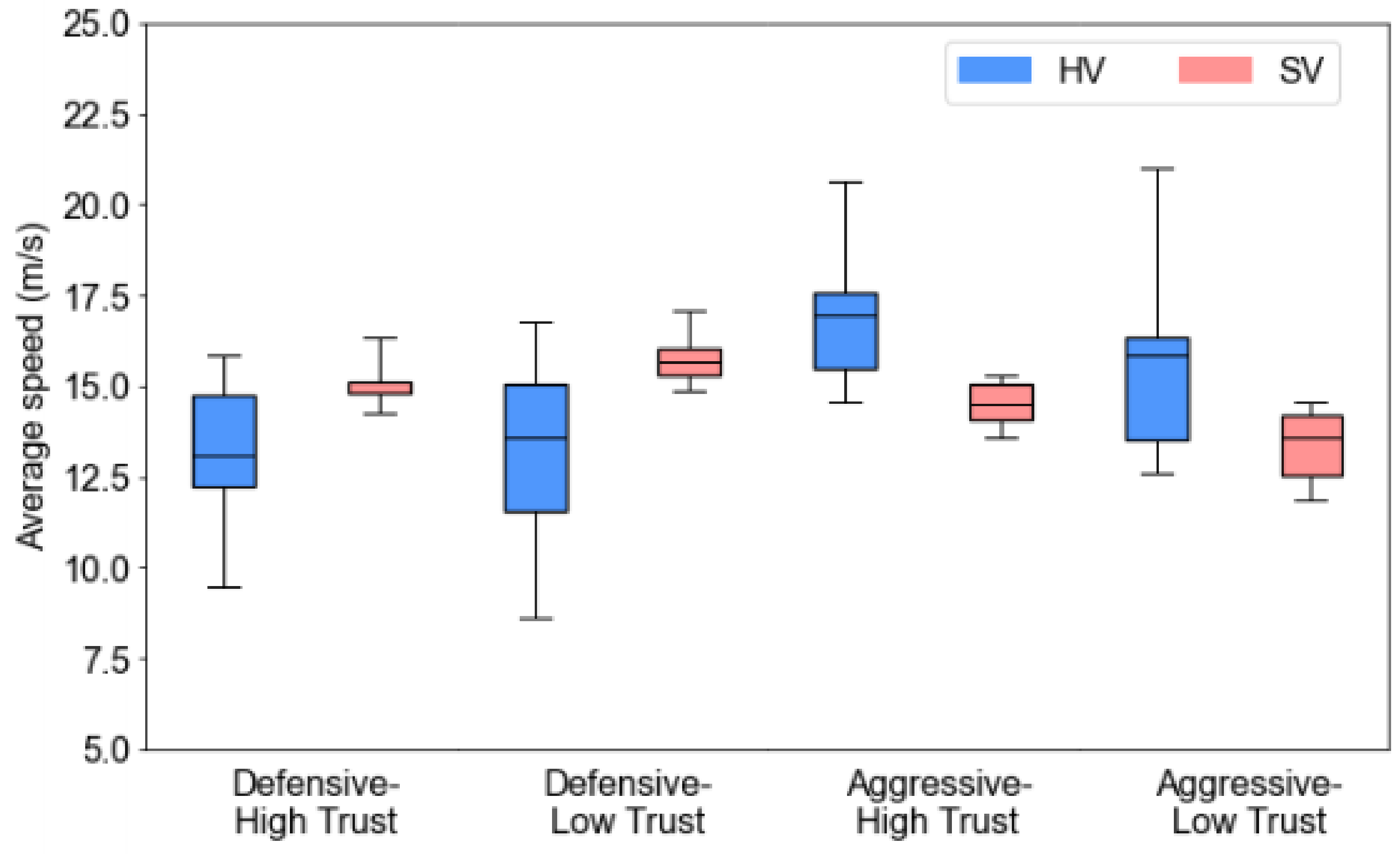

Fig. 5. Comparison of efficiency evaluation indicators

For **ride comfort** in Figure 6, D-H shows lower *jerk* values, indicating smoother rides. D-L increases jerk due to HV uncertainty. Aggressive drivers exhibit higher jerk overall, with A-L reaching peaks up to 2.47 $m/s^3$. These results indicate that aggressive and low-trust interactions have the most significant negative impact on ride comfort. Across all four scenarios, it is also observed that the maximum jerk of the SV is consistently lower than that of the HV, demonstrating the superior comfort control capability of AV.

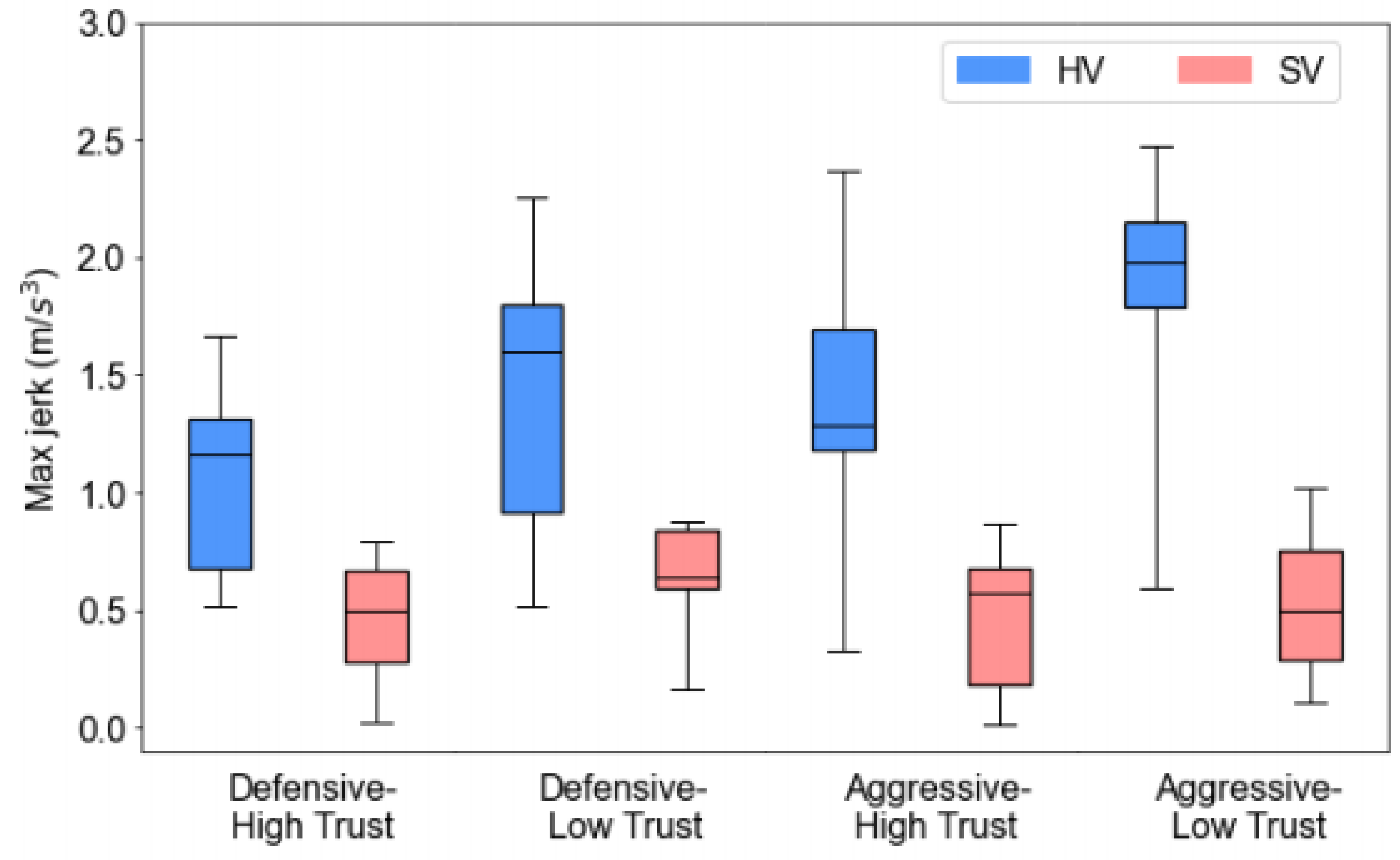

Fig. 6. Comparison of comfort evaluation indicators

### D. Ablation Study

An ablation study is conducted to assess the impact of the proposed trust mechanism. We compare the proposed TGLD, which integrates dynamic trust updates into the game-theoretic framework, against a baseline model without trust-awareness. Both models are tested under identical highway on-ramp merging simulations, and their performance is evaluated across safety, efficiency, and comfort using a paired-sample Wilcoxon signed-rank test.

The proposed TGLD shows a significant increase in average speed ($p<0.001$), improving efficiency, and a significant reduction in maximum jerk ($p=0.002$), enhancing ride comfort. No significant difference is found in minimum time headway ($p=0.157$), and all safety metrics remain within acceptable margins, with no collisions observed. As shown in Figure 7, the proposed TGLD achieves higher average speeds (13–15 m/s) compared to the baseline (11–13 m/s), reflecting more assertive and coordinated lane changes. Ride comfort also improves, with larger bubble sizes indicating smoother acceleration and fewer abrupt movements. In contrast, the baseline model shows lower average speeds, greater headway variability, and higher jerk values.

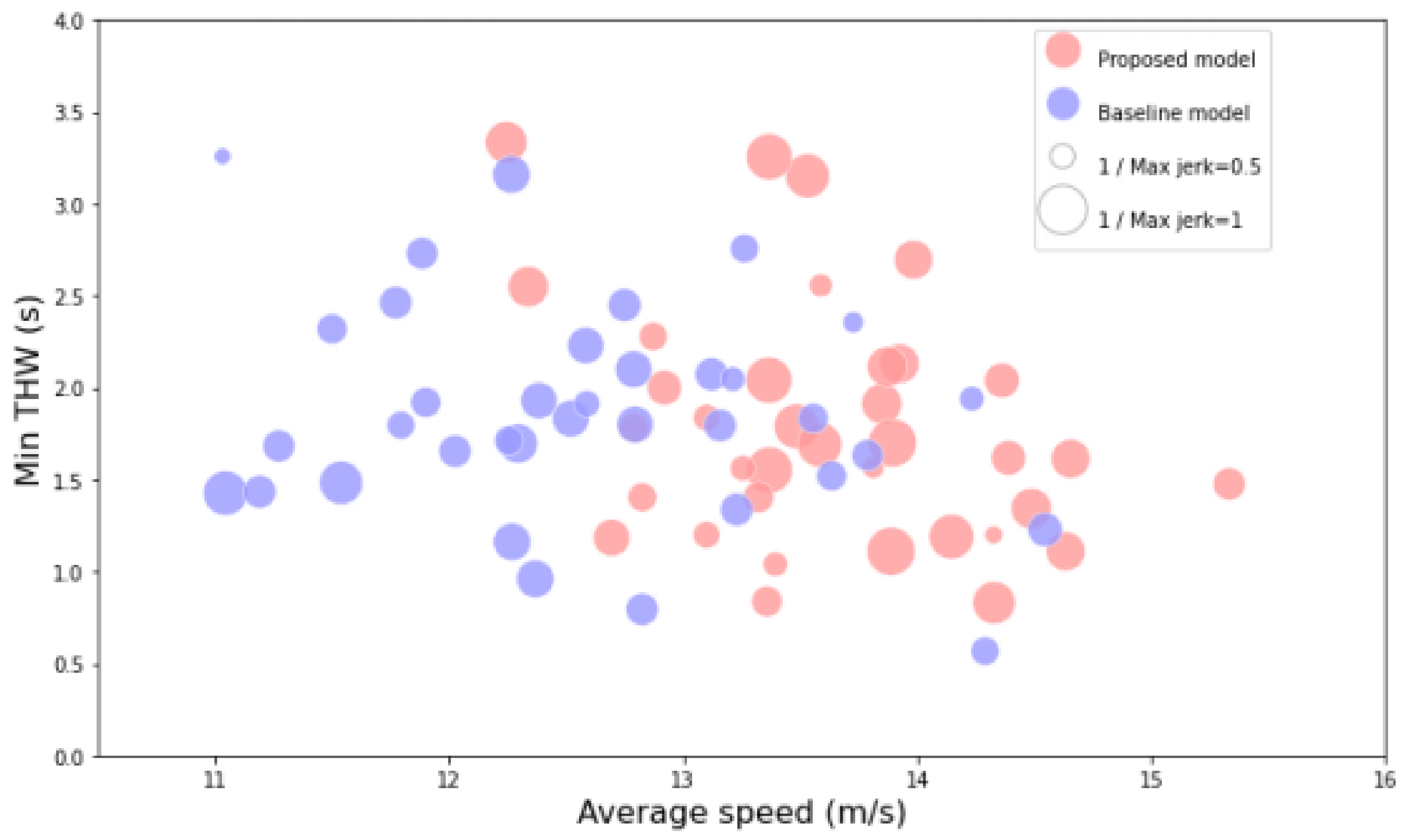

Fig. 7. Comparison of models with and without trust

## IV. CONCLUSIONS

In the context of heterogeneous traffic environments involving both HVs and AVs, this study proposes a TGLD framework. The proposed framework is validated in a highway on-ramp merging scenario through a real HIL simulation experiment involving participants with different driving styles. Experimental results demonstrate that low-trust human drivers tend to induce more defensive behaviors from AVs, resulting in greater speed differentials and decreased local traffic efficiency. In contrast, high-trust drivers support more predictable, smoother, and safe-preserving interactions. These observations highlight the importance of tailoring AV decision-making strategies to accommodate dynamic human driver trust levels. The core of the proposed TGLD framework is a dynamic trust estimation mechanism, which continuously updates the AV's assessment of the human driver's trust based on behavioral cues. Incorporating this trust mechanism into AV decision-making leads to higher average speeds and improved traffic efficiency, while still maintaining safety in heterogeneous traffic environments.